\documentclass[conference]{IEEEtran}
\IEEEoverridecommandlockouts

\usepackage{cite}
\usepackage{amsmath,amssymb,amsfonts}
\usepackage{algorithmic}
\usepackage{graphicx}
\usepackage{textcomp}
\usepackage{xcolor}
\usepackage{caption}
\usepackage{siunitx}
\usepackage{booktabs}
\usepackage{tabularx}
\usepackage{colortbl}
\usepackage[hyphens]{url}
\usepackage{breakurl}
\usepackage[gen]{eurosym}

\def\BibTeX{{\rm B\kern-.05em{\sc i\kern-.025em b}\kern-.08em
    T\kern-.1667em\lower.7ex\hbox{E}\kern-.125emX}}

\begin{document}

\title{Strong, Accurate, and Low-Cost Robot Manipulator}

\author{ }

\author{Georges Chebly$^{1}$, Spencer Little$^{1}$, Nisal Perera$^{2}$, Aliya Abedeen$^{2}$, Ken Suzuki$^{1}$, and Donghyun Kim$^{2}$
\thanks{$^{1}$ College of Engineering, University of Massachusetts Amherst, MA, U.S.}
\thanks{$^{2}$ College of Information and Computer Sciences, University of Massachusetts Amherst, MA, U.S. ({\tt\small donghyunkim@cs.umass.edu}) }}

\maketitle

\begin{abstract}
This paper presents \emph{Forte}, a fully 3D-printable, 6-DoF robotic arm designed to achieve near industrial-grade performance -- $0.63~\si{\kilogram}$ payload, $0.467~\si{\meter}$ reach, and sub-millimeter repeatability -- at a material cost under $\$ 215$. As an accessible robot for broad applications across classroom education to AI experiments, Forte pushes forward the performance limitations of existing low-cost educational arms. We introduce a cost-effective mechanical design that combines capstan-based cable drives, timing belts, simple tensioning mechanisms, and lightweight 3D-printed structures, along with topology optimization for structural stiffness. Through careful drivetrain engineering, we minimize backlash and maintain control fidelity without relying on high-power electronics or expensive manufacturing processes. Experimental validation demonstrates that Forte achieves high repeatability and load capacity, offering a compelling robotic platform for both classroom instruction and advanced robotics research.
\end{abstract}


\section{Introduction}
Can we build a 6-degree-of-freedom (DoF) robotic arm with a material cost under \$400, while achieving a half-meter workspace, a payload capacity of more than $0.5~\si{\kilogram}$, and repeatability within $0.5~\si{\milli\meter}$? We introduce Forte, a fully 3D-printed robotic manipulator, developed to affirmatively answer this question. In light of surging interest in robotics and artificial intelligence, providing accessible, hands-on educational tools has never been more important, as practical experience and experimental validation are essential components of robotics education. Despite a shared consensus on the need for accessible educational robots, most existing low-cost robotic arms exhibit limited capabilities (e.g., DoF, payload, reachability, control bandwidth), restricting their use to basic motion demonstrations.~\cite{UMIbot,cadene2024lerobot,kochrobot,hiwonder5dof,owi535}. 

These limitations make it challenging to explore advanced topics such as imitation learning or reinforcement learning for daily object grasping, complex bimanual manipulation, or dynamic object handling. Even a common household object can weigh over $300~\si{\gram}$, and with a gripper attached, a robot often needs to support $500~\si{\gram}$ or more. Moreover, many educational arms, such as such as Tiboni et al.’s low-cost 4-DoF educational manipulator~\cite{Tiboni2018} and Wlkata~\cite{WLKAtArobot}, suffer from a limited range of motion, making it difficult to design interactive or task-oriented experiments. Some high-performance arms do exist under \$600, such as RR1~\cite{RR1} and AR4~\cite{AR4}, but they often rely on large stepper motors that require high-power motor controllers, pushing total system costs beyond \$1000. Currently, there is no available robotic arm offering high performance at a low cost. 

\begin{figure}
\centering
\includegraphics[width=\linewidth]{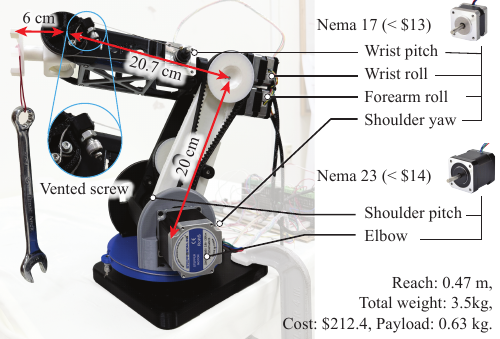}
\caption{{\bf Forte: A 6-DoF, 3D-printable robotic manipulator.} Built with low-cost components, Forte can lift a $0.63~\si{\kilogram}$ payload and achieve high-accuracy position control.}
\label{fig:Forte}
\end{figure}

Designing a manipulator that balances structural robustness, precision, and affordability presents fundamental challenges. The high manufacturing cost of metal parts makes aluminum structures impractical for low-volume educational use. Thus, we decided to develop a fully 3D-printable structure, making it feasible to duplicate the robot across classrooms or labs at a scale of 30–40 units. To minimize cost, we use metal components only for essential parts like bearings, screws, and motors -- all of which are inexpensive and commercially available. We also standardized motor and bearing selection to reduce component variability and enable bulk purchasing. To avoid high-power electronics, we used compact stepper motors (e.g., NEMA 17 and NEMA 23), each costing less than \$14. Although we have not yet developed a custom control board (which could further reduce cost), we successfully reduced the total system cost—including the off-the-shelf motor controllers used in this study—to under \$215, which is comparable to other open-source platforms.

\begin{table*}[!ht]
\vspace{-2mm}
    \centering
    \begin{tabular}{ m{3cm}  c c  c c  c c}
        Name & DoF & Reach ($\si{\meter}$) & Payload (\si{\kilogram}) & Material cost & Repeatability (\si{\milli\meter}) & Weight (\si{\kilogram})\\ [0.7 mm]
        \hline
         \toprule
       Umibot \cite{UMIbot}  & 5 & 0.3 & - & \$ 70 & - & -\\ [0.7mm]
        \hline
       Real Robot 1  \cite{RR1}  & 6 & 0.6 & 1 & \textgreater  \$500 $^\text{a}$  & - &  15 kg \\ [0.7mm] 
        \hline
       Worm drive \cite{howard2023design}  & 3 & $\sim 0.8$ & - & \pounds 478 & - & -\\ [0.7mm] 
        \hline
       Delta Z \cite{Deltaz}  & 3 & 0.06 & - & \$ 50 & 0.5 & -\\ [0.7mm] 
        \hline
       iArm \cite{zeng2022iarm}  & 6 & 0.37 & - & - & - & -\\ [0.7mm] 
        \hline
       Niryo Ned2 \cite{niryo_ned2_2025}  & 6 & 0.49 & 0.3 & \euro{} 3999 & 0.5 & 7\\ [0.7mm] 
        \hline
       Robotic Platform \cite{RoboticPlatform}  & 5 & $\sim 0.318$ & 0.1 & \euro{} 250 & 3 & -\\ [0.7mm] 
        \hline
       Wlkata Mirobot \cite{WLKAtArobot}  & 6 & 0.325 & 0.25 & \$ 1850 & 0.2 & 1.37\\ [0.7mm] 
        \hline
       Dobot Magician \cite{robotlab_dobot_2025}  & 4 & 0.32 & 0.5 & \$ 1999 & - & 4\\ [0.7mm] 
          \hline
       R-BL Project \cite{Tiboni2018}  & 4 & 0.6 & 0.2 & \euro{}500  & 0.5 &  \textgreater 8 $^\text{c}$ \\ [0.7mm]
       \hline
       Kinova Gen3 Lite \cite{kinovagen3lite}  & 6 & 0.76 & 0.5 & \$ 13500 $^\text{b}$ & - & 5.4\\ [0.7mm]
               \hline
                  \toprule
        Forte (ours) & 6 & 0.467 & 0.63 & \$ 212.4 & 0.467 & 3.5 \\[0.7mm] 
        \toprule
    \end{tabular}

    \begin{minipage}{\textwidth}
    \footnotesize
    \textsuperscript{a} Estimated price based on the motors and electronics used.
     \textsuperscript{b} Minimum Price on most selling websites 
    \textsuperscript{c} Weight estimated based on material, bearings, and motors
    \end{minipage}
    \caption{Specifications of small-scale robot arms.}
    \label{tb:comparison_robot}

\end{table*}
 
Designing a robot that can lift more than $500~\si{\gram}$ and achieve $0.5~\si{\milli\meter}$ repeatability with a $0.5~\si{\meter}$ reach at this price point presents unique challenges. While torque amplification via speed reduction is one solution, achieving high reduction ratios often compromises cost, control bandwidth, or both. Conventional gear mechanisms (e.g., planetary, harmonic, cycloidal drives) require precision manufacturing and solid materials, which introduce friction, backlash, and compliance, all detrimental to control performance \cite{garcia2020compact}. In our design, backlash and drivetrain stiffness were critical concerns, as they directly affect repeatability and control bandwidth.

To address these challenges, we employed a steel cable-based capstan mechanism combined with a timing belt system for the shoulder and elbow joints. We also introduced a simple yet effective cable-tensioning method using vented screws \cite{mcmaster_carr_nodate}, enabling easy assembly and maintenance. While capstan drives offer low backlash and high torque transmission, applying them to every joint would unnecessarily complicate the design, increase weight, and raise the component count without significant performance gains. Therefore, for the wrist and upper arm roll joints -- where the mechanical load is lighter and backlash has less influence -- we opted for conventional gears to simplify the drivetrain. Additionally, we placed three motors at the elbow to counterbalance their weight and reduce the torque burden of the elbow motor. This careful design allowed us to achieve a $0.63~\si{\kilogram}$ payload and $0.5~\si{\milli\meter}$ repeatability -- performance levels comparable to some industrial arms of similar size~\cite{kinovagen3lite,meca500,motomini} and unmatched among current open-source designs \cite{niryo_ned2_2025,RoboticPlatform}(see Table.~\ref{tb:comparison_robot}).

In addition to the efforts to enhance the drivetrains, we also aimed to optimize linkage design to maximize structural stiffness while minimizing mass. The result is a lightweight arm structure (only $0.18~\si{\kilogram}$ for upperarm and $0.11~\si{\kilogram}$ for forearm), with the total system weighing approximately $3.5~\si{\kilogram}$ -- mostly attributable to the motors. Combined with a stiff cable-belt transmission system, Forte exhibits minimal structural oscillation or vibration, enhancing its practical usability.

Forte offers a combination of low cost, mechanical performance, and control fidelity. The contributions of this work are threefold: 1) development of a robotic arm with high control fidelity under tight budget constraints, 2) cost-effective mechanical mechanisms such as -- the cable tensioner, belt-cable hybrid transmission, and a differential gear at the wrist  -- to enhance performance, and 3) experimental validation of the system, demonstrating compelling repeatability and payload capacity that suggest real potential to resolve the longstanding dilemma between cost and performance in robotic arms.

\section{Survey of education-purpose robot arms}
Even entry-level systems such as the KUKA youBot are beyond the reach of most teaching labs and small enterprises~\cite{bischoff2011kuka}. Their heavy weight, rigid designs and end‑effector tooling offer desirable accuracy and payload, but these systems remain prohibitively expensive and overly complex for broader applications such as educational labs or small‑scale prototype development.

Existing educational arms often force a trade‑off between cost and capability. Kits like LEGO Mindstorms offer an affordable introduction for 10–14‑year‑olds, but their plastic construction and limited stiffness make them ill‑suited for larger payloads or more demanding tasks \cite{mindell2000lego}. 
At the other end of the spectrum, desktop arms that do offer industrial‑style rigidity and precision, such as the Dobot Magician, WLKATA Mirobot, or the Niryo Ned2—come with $1,999+$USD price tags, putting them out of reach for many learners\cite{robotlab_dobot_2025, WLKAtArobot, niryo_ned2_2025}.

Zeng \emph{et al.} attempt to bridge this gap with the iArm, an inexpensive 6‑DOF kit designed for AI and robotics education \cite{zeng2022iarm}. Despite its promise, the iArm’s short reach, use of lower‑torque serial servos, and lack of mechanical leverage limit both its workspace and its payload capacity, constraining its utility in realistic manipulation and grasping exercises.

Likewise, recent open‑hardware efforts reveal complementary drawbacks. Real Robot One (RR1) brings stepper‑driven, 6‑DOF kinematics to the desktop, yet its heavy printed structure and large footprint constrain portability and its use in space‑constrained labs \cite{RR1}. On the other side, the UMIRobot kit ships worldwide for undergraduate teleoperation courses, but its five‑DOF layout lacks a wrist‑roll joint, and its short reach confines tasks to lightweight, close‑in manipulation\cite{UMIbot}. Academic prototypes such as Howard’s 3‑DOF worm‑drive arm were developed to offer a low‑cost planar manipulator for simple pick‑and‑place and agricultural demonstration tasks. This design achieves a roughly $0.45~\si{\meter}$ radial reach and supports up to 1 kg of static payload for under USD 300, yet its absence of wrist articulation and limited vertical workspace constrain its applicability to more complex three‑dimensional manipulation \cite{howard2023design}. Similarly, Patil \emph{et al.}’s DeltaZ delta‑style robot was designed to democratize real‑robot benchmarks in reinforcement‑learning labs, leveraging three translational DOFs and lightweight micro‑servos to operate within a 60 mm diameter workspace \cite{Deltaz}. Although DeltaZ excels in low‑inertia, high‑speed motion for small-scale tasks, its sub‑100 g payload and purely translational kinematics limit its use in broad manipulation tests. 

Tiboni \emph{et al.} guided teams through the end‑to‑end design of a 4‑DOF serial manipulator—complete with Denavit–Hartenberg–defined kinematics, belt‑driven DC motors with encoder feedback, and an FDM‑printed three‑finger gripper reinforced with elastic SBR fingertips—using a modular Raspberry Pi/Arduino control architecture~\cite{Tiboni2018}. Their prototype achieved a $60~\si{\centi\meter}$ reachable workspace and $0.5~\si{\milli\meter}$ repeatability, all for roughly €450 in parts. However, the robot's DoF is 4, and the payload is limited to $0.2~\si{\kilogram}$.

Meanwhile, Twist Snake, a 7-DOF cable-driven manipulator made from laser-cut acrylic and 3D printed components, incorporates the Dynamixel MX-106 motors, which costs more than \$600 per motor, which is expensive \cite{twist_snake_icra}.

Compared to existing low-cost robotic arms, our 6-DoF arm, Forte, delivers high repeatability and payload capacity while maintaining a low material cost. The entire structure is fabricated from 3D-printed PLA and assembled using a standardized set of bearings, belts, and sprockets, enabling streamlined and reproducible manufacturing. We envision its widespread duplication in educational settings and research labs, enabling users to explore topics ranging from simple kinematics studies to machine learning-based manipulation research.


\section{Mechanical design}
In designing Forte, we applied key engineering principles to maximize payload capacity and control performance while minimizing cost and complexity. A central design priority was reducing the weight borne by the structure and actuators, particularly at the distal segments of the arm. To achieve this, we mounted the stepper motors, the heaviest components, as close to the shoulder joint as possible. For the elbow actuation, we strategically positioned the motors and joints (elbow, forearm, and wrist) near the elbow itself to avoid an excessively long and complex drivetrain. Although this places some actuators farther from the shoulder, we compensated by minimizing torque demands at the elbow joint by counter-balancing three motors around the elbow joint.

By relocating actuators proximally and incorporating high-reduction mechanisms -- including capstan drives, timing belts, and gear trains -- we were able to both amplify available motor torque and enhance angular resolution with minimal backlash and friction. In this section, we describe the mechanical architecture of Forte in detail, highlighting the key design features that enable its high performance and accessibility.



\subsection{Capstan drive}

A capstan drive is a compact, high‑efficiency cable‑driven transmission commonly found in robotic manipulators \cite{kim2017anthropomorphic}, legged robots~\cite{hwangbo2018cable}, and haptic interfaces~\cite{dovat2006haptic,ivimey2025comparative}. By routing cables in a figure‑eight pattern around an input sheave and an output pulley, it delivers exceptionally low backlash and superior torque transmission efficiency \cite{mazumdar2017synthetic}.

\begin{figure}[t]
    \centering
    \includegraphics[width=\linewidth]{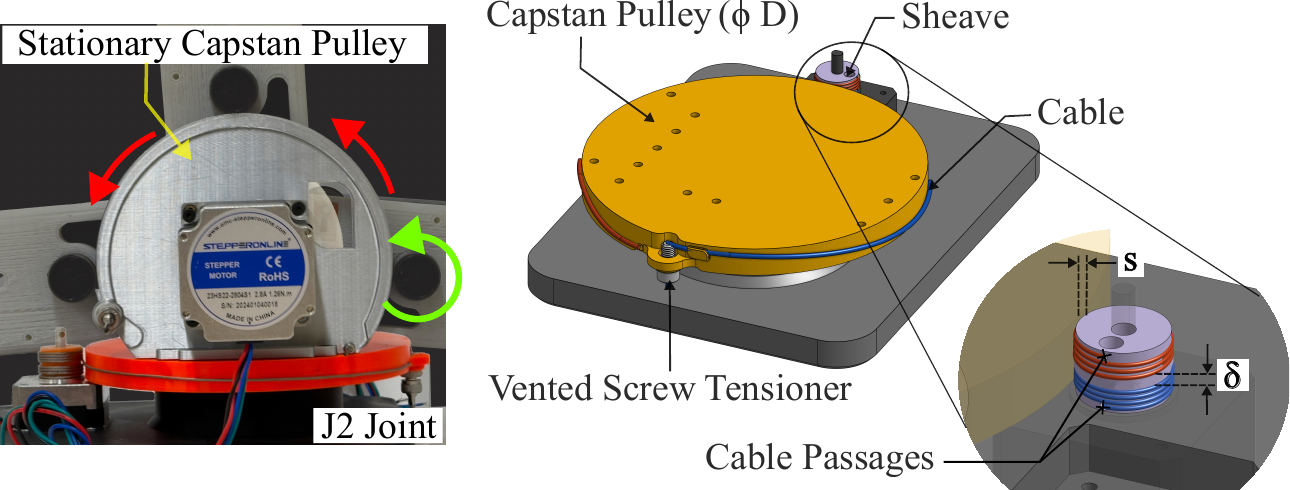}
    \caption{\textbf{Exploded view of the capstan drive at the base joint.} The large reduction ratio from the capstan pulley effectively increases the coarse resolution of the stepper motor. }
    \label{fig:figure2}
\end{figure}

For Forte, a capstan mechanism was implemented at the first shoulder yaw joint ($J_1$) and at the second shoulder pitch joint ($J_2$). The main advantage of a capstan mechanism is its ability to enable a cost-effective transmission system, requiring only a single cable and two pulleys, minimal friction, and high stiffness if engineered properly.

\begin{table*}[ht]
\centering
\begin{tabular}{|c|c|c|c|c|c|}
\hline 
\textbf{Joint} & \textbf{Motor} & \textbf{Holding Torque (Nm)} & \textbf{Mechanism} & \textbf{Amplification} & \textbf{Max Joint Torque (Nm)} \\
\hline
1 (Shoulder Yaw) & Nema 17 & 0.157 & Capstan Drive & 8:1 & 1.256 \\
2 (Shoulder Pitch) & Nema 23 & 1.26 & Belt + Capstan & 18.75:1 & 23.625 \\
3 (Elbow Pitch) & Nema 23 & 1.26 & Belt Transmission & 2.5:1 & 3.15 \\
4 (Forearm Roll) & Nema 17 & 0.157 & Gear Drive & 3.33:1 & 0.52281 \\
5 (Wrist Pitch) & Nema 17 & 0.157 & Cable Drive & 3:1 & 0.471 \\
6 (Wrist Roll) & Nema 17 & 0.157 & Gear Drive & 3.33:1 & 1.256 \\
\hline
\end{tabular}
\caption{Motor and torque specifications for each joint}
\label{tb:torque_specs}
\end{table*}

The drive consists of a sheave of diameter $D_0 = 19.4\,\text{mm}$ and a pulley of diameter $D = 155.2\,\text{mm}$, which offers an 8:1  reduction ratio. The sheave requires eight full windings of cable to provide approximately $360^\circ$ of motion for the capstan pulley, which serves as the base for the entire arm. To ensure a smooth and compact configuration, the height $h$ of the sheave and pulley must be matched to accommodate all windings. This height is determined by the cable thickness $t$ and the transmission ratio $\gamma$:

\begin{equation}
    h = t \gamma + \delta,
    \label{eq:height}
\end{equation}
where $\delta$ is a design tolerance introduced to prevent excessive contact between windings. For this application, a tolerance of $\delta = 2\,\text{mm}$ is selected to reduce the risk of noise, energy dissipation, and premature wear due to friction between adjacent cable layers.
Additionally, the ideal lateral spacing between the sheave and pulley is given by:

\begin{equation}
    s = 1.5 t
    \label{eq:spacing}
\end{equation}

The cable is routed through a series of through-holes in a custom 3D-printed sheave to ensure proper fixation and routing integrity, as illustrated in Fig.~\ref{fig:figure2}(a). The red segments represent the top wiring, which is clamped from the bottom on the capstan pulley side. The blue segments begin at the bottom of the sheave and are clamped on the top side of the capstan. Note that both the red and blue segments belong to a single continuous cable that passes through the interior of the sheave.

For instance, if the sheave directly connected to the motor rotates counterclockwise, the red segments shown in Fig.~\ref{fig:figure2} (a) wrap around the sheave, while the blue segments unwind as the capstan rotates clockwise. To ensure smooth operation and mechanical reliability, proper cable tension is critical. In Forte’s capstan system, this is achieved using a vented through-hole screw (see Fig.~\ref{fig:Forte}). The steel cable is inserted through a screw and secured with a ball fitting. Cable tension is adjusted using a screw, which is then locked in place with a nut.

A similar design concept is employed at the ($J_2$) joint. However, in this case, the capstan pulley remains stationary while the sheave -- mounted to one of the upper arm plates -- rotates freely as it moves around the capstan. The cable wrapping and unwrapping follow the same principle described earlier in this section. As the cable is pulled, it causes the sheave to rotate around the stationary capstan, thereby enabling the pitch motion of the entire arm as shown in Fig.~\ref{fig:figure2} (a). An important difference between $J_1$ and $J_2$ joints is the calculation of the \textit{reduction ratio}. In case of the shoulder pitch joint, the reduction ratio, $\gamma$, is given by:

\begin{equation}
    \gamma = \frac{D}{D_0} + 1
    \label{eq:reduction}
\end{equation}

This formulation arises because the capstan (large pulley) is stationary, while the smaller pulley not only rotates about its own axis, but also revolves around the stationary capstan. This revolution introduces an additional full rotation of the small pulley due to the circular path it traces. In other words, as the small pulley revolves once around the stationary capstan, it gains an extra rotation equivalent to traveling a circular path of circumference  \( 2\pi D_0 \). Using $D_0 = 130~\si{\milli\meter}$ and $D=20~\si{\milli\meter}$, the capstan yields a reduction ratio of 7.5. Combined with a 2.5:1 reduction from the belt drive, the total reduction ratio at the shoulder pitch joint is 18.75. It is worth noting that the belt transmission employs a Gates PowerGrip GT3 5MGT-9 belt \cite{gates_belt}, which was chosen for its high stiffness to enable high-bandwidth control.

\begin{figure}
    \centering
    \includegraphics[width=1\linewidth]{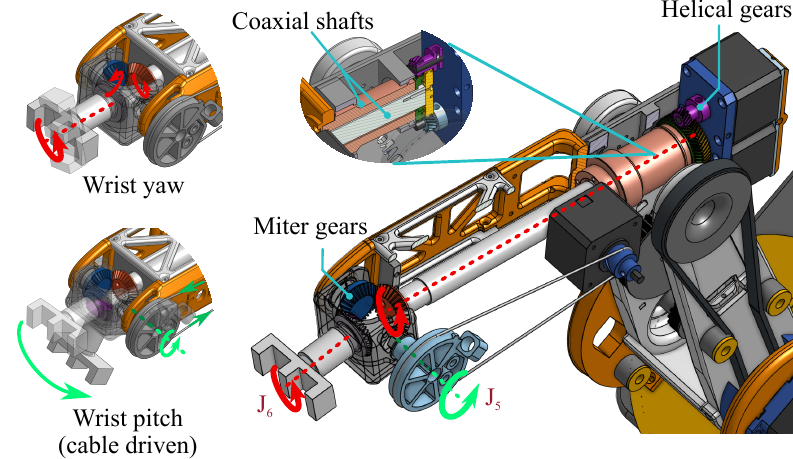}
    \caption{\textbf{Degrees of freedom in the wrist joint.} A coaxial shaft placed along the forearm drives a series of miter gears, rotating the wrist yaw joint. A cable drive rotates the wrist pitch joint. }
    \label{fig:wrist-dof}
\end{figure}

\subsection{Wrist mechanism}
As illustrated in Fig. \ref{fig:wrist-dof}, the wrist features three degrees of freedom in a roll--pitch--roll configuration. Two stepper motors mounted near the elbow joint handle the roll motions of both the forearm and the end effector. The wrist pitch joint is actuated via a cable-driven mechanism. Its stepper motor is also located near the elbow joint, counterbalancing the other two motors to reduce the elbow joint torque needed to support the forearm.

The first two motors (i.e., for forearm roll and wrist roll) are geared with a 3.33:1 reduction ratio and drive coaxial shafts. For this gear set, helical gears with a pressure angle of \(20^{\circ}\) and a helix angle of \(30^{\circ}\) were selected over conventional spur gears. Testing showed that 3D-printed helical gears produced less vibration due to smoother tooth engagement, consistent with~\cite{flodin2000wear}, which found spur gears wear faster due to abrupt contact transitions.

The inner shaft extends through the forearm to drive the end effector via a bevel gear system, while the outer shaft rotates the entire forearm. To ensure stability and alignment within the bevel gear system, the inner shaft is supported by three bearings, which help prevent slippage during operation. A summary of Forte’s joint configuration and specifications is provided in Table~\ref{tb:torque_specs}.

\subsection{Topology Optimization} \label{sec:topOpt}

One of Forte’s key features is its ability to lift payloads of up to $0.63~\si{\kilogram}$ while utilizing low-cost motors. In addition to the effort to increase joint torque, the robot's structure must also satisfy requirements for both \textit{affordability} and \textit{structural integrity}, as the arm must not fail under operational loads. In other words, it must withstand the expected forces without yielding or breaking, even when fabricated from \textit{cost-effective materials}. Forte is fabricated from PLA (Polylactic Acid) due to its low cost, printability, and lightweight properties. Although PLA lacks high mechanical strength, its use is justified through structural analysis and optimization. Stress analysis ensures reliability under load, while topology optimization prevents overdesign, reducing mass, cost, and potential impacts on dynamic performance.

To verify the arm’s durability, a finite element analysis (FEA) was conducted in \textsc{Ansys} using PLA material properties reported by M. Rismalia \textit{et al.}~\cite{rismalia2019infill}, including tensile yield strength, ultimate tensile strength, and Young’s modulus. Since values were only provided for 25\% and 50\% infill densities, linear interpolation was used to estimate the properties at a 30\% infill. The interpolated values were: Young’s modulus \(E = 2.768\,\text{GPa}\), yield strength \(\sigma_y = 28.4\,\text{MPa}\), and ultimate tensile strength \(\sigma_u = 30.06\,\text{MPa}\).

\begin{figure}[t]
    \centering
    \includegraphics[width=\linewidth]{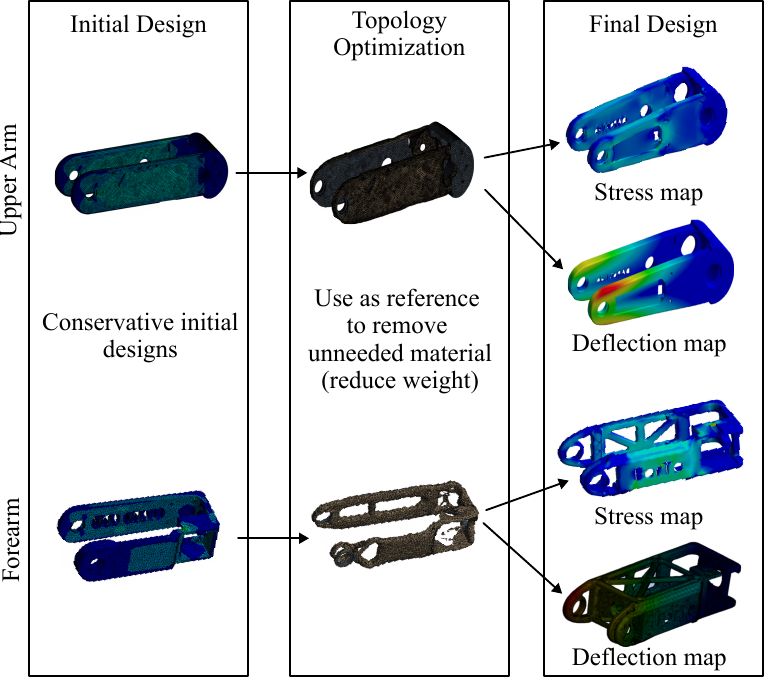}
    \caption{\textbf{Topology optimization process for the upper arm and forearm.} Conducted in Ansys Workbench$^{\text{\textregistered}}$, the topology optimization process helps reduce weight of the structure, while improving stiffness.}
    \label{fig:topOpt}
\end{figure}

Our topology optimization process followed a methodology similar to \cite{perera2024staccatoe}. The simulation accounted for torsional and payload forces, as well as the self-weight of the motors and printed material, to reflect realistic operating loads on the initial upper arm and forearm designs, which were the focus due to their role as the main load-bearing structures. As shown in Fig.~\ref{fig:topOpt}, topology optimization was then applied to reduce structural mass without compromising load-bearing performance, starting from a conservative baseline design. This optimization led to significant weight reductions in both the upper arm and forearm, as detailed below:
 
\begin{itemize}
    \item The upper arm’s weight was reduced from \textbf{0.337 kg} to \textbf{0.18075 kg}, retaining approximately \textbf{53\%} of the original mass.  
    \item The forearm retained \textbf{63.88\%} of its original mass of \textbf{0.1108 kg} ending up with a \textbf{0.069 kg} weight.
\end{itemize}

A final static structural analysis verified the integrity of the optimized designs, yielding a factor of safety (FoS) of 4.38 for the upper arm and 1.57 for the forearm. A higher FOS in the upper arm is preferred due to its greater exposure to loads and moments. To mitigate deformation under torsional loads up to $1~\si{\newton\meter}$, lightweight torsional supports were added to the forearm, significantly improving stiffness and ensuring positional stability under combined loading conditions.





\begin{figure}[!t]
    \centering
    \includegraphics[width=\linewidth]{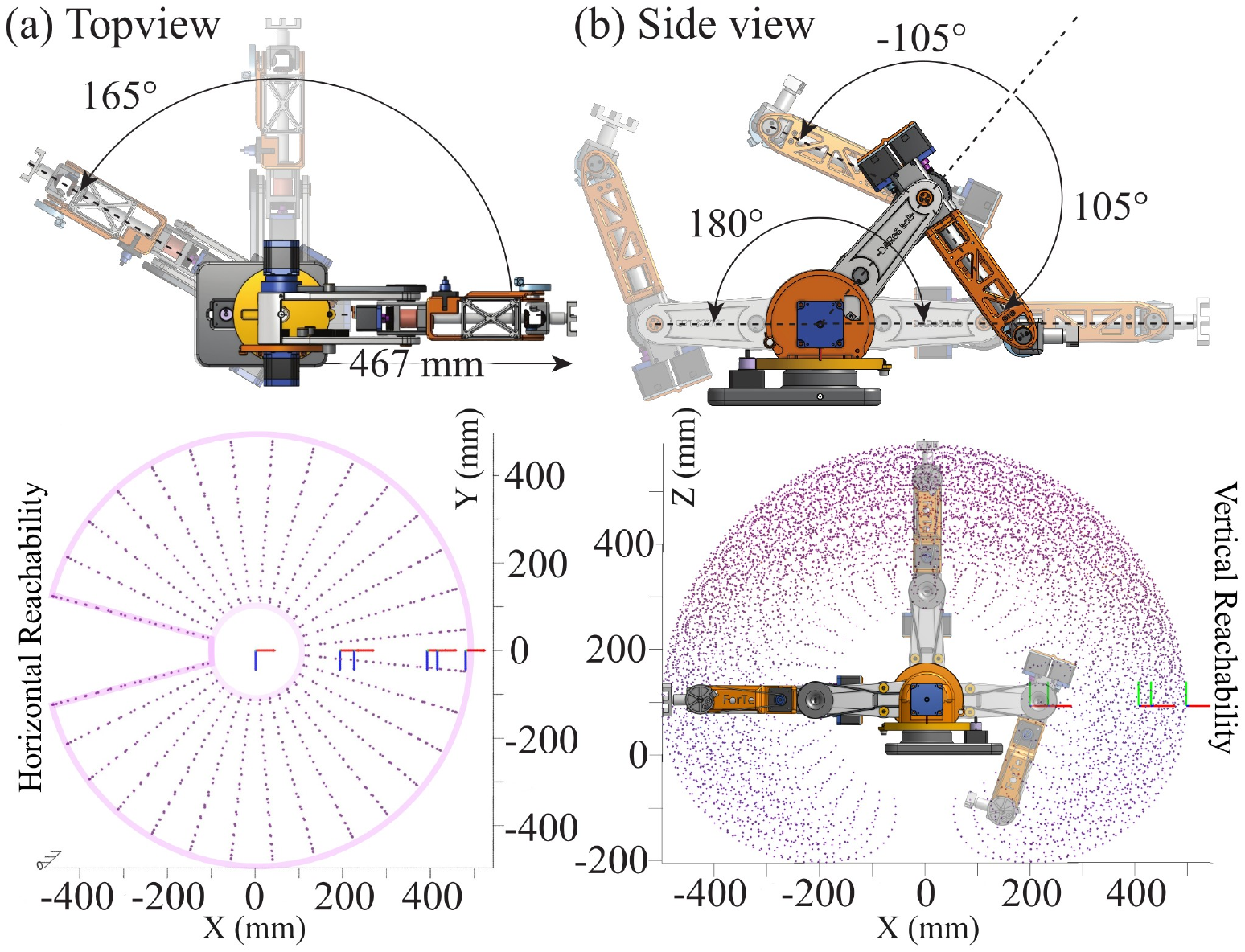}
    \caption{{\bf Range of motion.} Forte offers a large range of motion and workspace thanks to the carefully designed linkage structure to avoid collisions between linkages.}
    \label{fig:workspace}
\end{figure}

\subsection{Workspace}

Fig.~\ref{fig:workspace} provides an overview of the robot’s workspace along with each joint's range of motion. When fully extended, the robot can reach up to 467~mm from its base, as shown in the top view (Fig.~\ref{fig:workspace}a). The shoulder (base) joint employs a capstan mechanism that enables a wide rotational range of 330$^{\circ}$, exceeding the typical 180$^{\circ}$–270$^{\circ}$ limits of most capstan-driven systems \cite{ridgway_towards_nodate}, \cite{tomaszewski_designing_nodate}.

In the side view (Fig.~\ref{fig:workspace}b), the vertical joint configuration is illustrated. The upper arm and forearm achieve a combined pitch motion of 210$^{\circ}$ (from $-105^\circ$ to $+105^\circ$), and the elbow enables an additional 180$^{\circ}$ of articulation. This joint architecture allows Forte to access points beneath its own base, which is critical for operations requiring high dexterity in constrained environments.

The XY and XZ workspace plots, derived from the Denavit–Hartenberg (DH) parameters listed in Table~\ref{tab:dh_params}, confirm Forte’s capability to cover a wide area both horizontally and vertically. The presence of three independent pitch joints at different elevations further enhances the robot's ability to manipulate in 3D space, demonstrating both reach and maneuverability uncommon in comparable low-cost robotic arms.

\begin{table}[ht]
\centering
\begin{tabular}{|c|c|c|c|c|}
\hline
\textbf{Link} & \boldmath$\theta_i$ & \boldmath$\alpha_{i-1}$ & \boldmath$a_{i-1}$ (mm) & \boldmath$d_i$ (mm) \\
\hline
1 & $\theta_1$ & $\frac{\pi}{2}$ & 0 & 93.53312 \\
2 & $\theta_2$ & $0$             & 200  & 0 \\
3 & $\theta_3$ & $\frac{\pi}{2}$ & 34 & 0 \\
4 & $\theta_4$ & $-\frac{\pi}{2}$ & 173 & 0 \\
5 & $\theta_5$ & $\frac{\pi}{2}$ & 23.88 & 0 \\
6 & $\theta_6$ & $0$             & 67.45 & 0 \\
\hline
\end{tabular}
\caption{DH Parameters of Forte}
\label{tab:dh_params}
\end{table}


\section{Material cost}
A comprehensive bill of materials was prepared to demonstrate the affordability of the ForTe arm. As shown in Table~\ref{tab:Forte_bom}, the total cost for one arm---including motors---amounts to less than \$215. This makes ForTe a highly cost-effective and accessible alternative compared to other robotic arms on the market, as presented in Table~\ref{tb:comparison_robot}.

The table reflects the cost of producing 25 arms, with components ordered in bulk to take advantage of discounted pricing. Prices were calculated precisely, accounting for every screw and nut used. Additionally, the costs of raw materials and steel cables were carefully estimated.

To determine material usage, all 3D-printed parts for ForTe were imported into \texttt{PrusaSlicer} with supports enabled. The total weight of PLA filament required to print one arm was 1080.21 grams, factoring in a 30\% infill for the forearm and upper arm, as modeled in the Ansys simulation discussed in Section~\ref{sec:topOpt}. This number was then multiplied by 25 to estimate the total number of 1\,kg spools needed. 

For the steel cables, ForTe requires lengths of 1100\,mm, 700\,mm, and 400\,mm for the shoulder yaw, shoulder pitch, and wrist pitch, respectively. A single spool of 55{,}000\,mm (181\,ft) can be sourced from most cable suppliers.

Every detail has been meticulously accounted for in this bill of materials. The final total was divided by 25 to obtain the per-unit cost of one arm.

\begin{table}[ht]
\centering
\scriptsize
\renewcommand{\arraystretch}{1.05}
\setlength{\tabcolsep}{3.5pt}

\definecolor{HeaderBlue}{HTML}{D6EAF8}

\begin{tabularx}{\columnwidth}{@{}lXrrr@{}}
\toprule
\textbf{Category} & \textbf{Item} & \textbf{Qty} & \textbf{Unit (\$)} & \textbf{Total (\$)} \\
\midrule

\rowcolor{HeaderBlue}
\multicolumn{5}{l}{\textbf{Mechanical Components}} \\
& Nema 17                        & 100 & 11.20  & 1120.00 \\
& Nema 23                        & 50  & 11.65  & 582.50 \\
& 62x50x6 mm bearings            & 100 & 2.763  & 276.30 \\
& 21x15x4 mm bearings            & 275 & 0.999  & 274.725 \\
& 15x10x4 mm bearings            & 200 & 0.9899 & 197.98 \\
& 37x30x4 mm bearings            & 50  & 1.89   & 94.50 \\
& Power grip GT3 300-5MGT-9     & 25  & 14.00  & 350.00 \\
& Power grip GT3 540-5MGT-9     & 25  & 19.00  & 475.00 \\
& Steel cables (ft)             & 181.25 & 3.11 & 563.6875 \\
& Cable fittings                & 75  & 1.68   & 126.00 \\
& M6 vented screw               & 75  & 2.40   & 180.00 \\

\textit{M3 Countersink Screws} & 35 mm length      & 100  & 0.15   & 15.00 \\
& 30 mm length                  & 100  & 0.07   & 7.00 \\
& 10 mm length                  & 375  & 0.07   & 26.25 \\

\textit{M3 Counterbore Screws} & 35 mm length      & 200  & 0.18   & 36.00 \\
& 20 mm length                  & 100  & 0.09   & 9.00 \\
& 10 mm length                  & 325  & 0.09   & 29.25 \\

\textit{M2.5 Counterbore Screws} & 12 mm length     & 125  & 0.0788 & 9.85 \\

\textit{M2 Counterbore Screws} & 20 mm length       & 200  & 0.08   & 16.00 \\
& 12 mm length                  & 550  & 0.08   & 44.00 \\

\textit{Nuts} & M3 Nut                          & 100  & 0.0459 & 4.59 \\
& M6 Nut                          & 75   & 0.0699 & 5.2425 \\

\textit{Miscellaneous} & M3 Set Screws           & 150  & 0.06   & 9.00 \\
& Washers                       & 75   & 0.02   & 1.50 \\
& M3 Heat Inserts               & 1200 & 0.03795 & 45.54 \\
& M2.5 Heat Inserts             & 125  & 0.02495 & 3.11875 \\
& M2 Heat Inserts               & 750  & 0.071 & 53.25 \\

\rowcolor{HeaderBlue}
\multicolumn{5}{l}{\textbf{Material}} \\
& PLA 1 kg spool                & 27   & 9.99   & 269.73 \\

\rowcolor{HeaderBlue}
\multicolumn{5}{l}{\textbf{Electrical Components}} \\
& A4988 Motor Driver            & 100  & 0.71   & 71.00 \\
& TB6600 Stepper Driver         & 50   & 7.35   & 367.50 \\
& Arduino Uno                   & 25   & 1.86   & 46.50 \\

\midrule
\multicolumn{4}{r}{\textbf{Total for 25 Arms:}} & \textbf{5310.013} \\
\multicolumn{4}{r}{\textbf{Total per Arm:}}     & \textbf{212.40} \\
\bottomrule
\end{tabularx}
\caption{Bill of materials for Forte arm fabrication (batch of 25 arms)}
\label{tab:Forte_bom}
\end{table}

\section{Experiments}
To experimentally validate the repeatability and maximum payload, we conducted the real-hardware tests involving repeatable motions with and without payload. Fig.~\ref{fig:exp1}(a) shows the experimental setup. Stepper motors were controlled using an Arduino UNO with TB6600 and A4988 stepper motor drivers, which are affordable and widely available, and powered by $24$V DC supplies.

\subsection{Repeatability Analysis}
To evaluate the repeatability and precision of our low-cost robotic arm, we repeatedly articulated it to a reference position and measured deviations using a Westward 4KU81 dial indicator over multiple motion cycles. To achieve high positional consistency, we used the built-in microstepping capability of the TB6600 stepper driver, which significantly reduces step-induced oscillations and improves fine-motion control. Experiments were conducted at linearly increasing motor speeds, starting from $500$ steps/s and incrementing by $500$ steps/s up to $2500$ steps/s. For each speed, $5$ to $10$ motion cycles were recorded. 

\begin{figure}[!t]
    \centering
    \includegraphics[width=\linewidth]{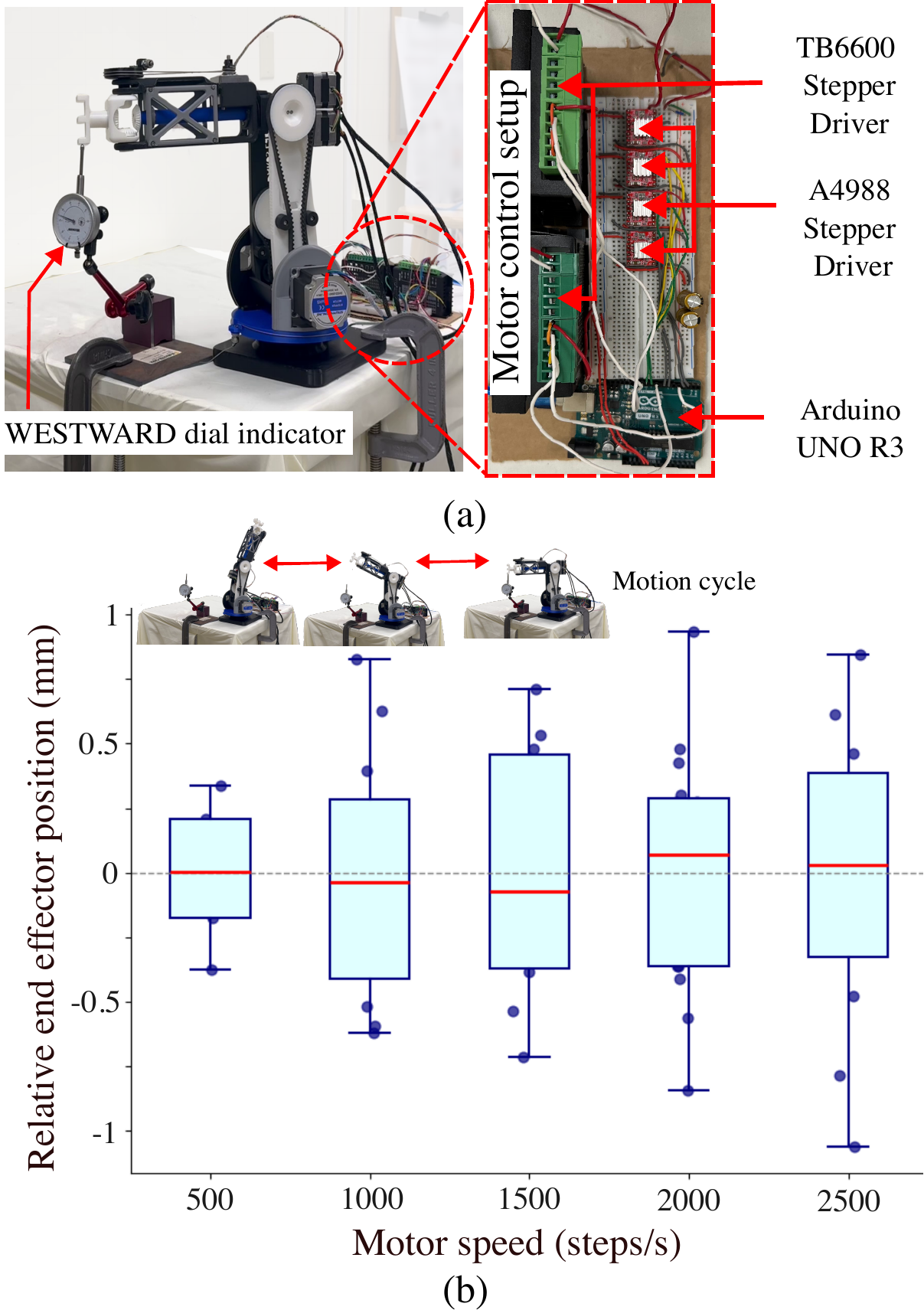}
    \caption{{\bf Repeatability analysis.} (a) The experiment setup with a dial indicator to validate reliable position control. (b) Box-plot representing results from repeatability experiments, displaying $\leq 0.6~\si{\milli\meter}$ precision.}
    \label{fig:exp1}
\end{figure}

As shown in Fig.~\ref{fig:exp1}b, the robotic arm demonstrated sub-millimeter repeatability across all tested speeds. The highest standard deviation of $0.587$~\si{\milli\meter} occurred at the maximum speed of $2500$ steps/s, while the lowest deviation of $0.286$$~\si{\milli\meter}$ was recorded at $500$ steps/s. Overall, the results indicate a trend of increasing positional deviation with speed. The average deviation across all tested speeds was $0.467~\si{\milli\meter}$. Despite its low-cost construction, Forte achieves repeatability comparable to significantly more expensive robotic arms, as highlighted previously in Table~\ref{tb:comparison_robot}.

\subsection{Maximum payload}
\begin{figure}[t]
    \centering
    \includegraphics[width=\linewidth]{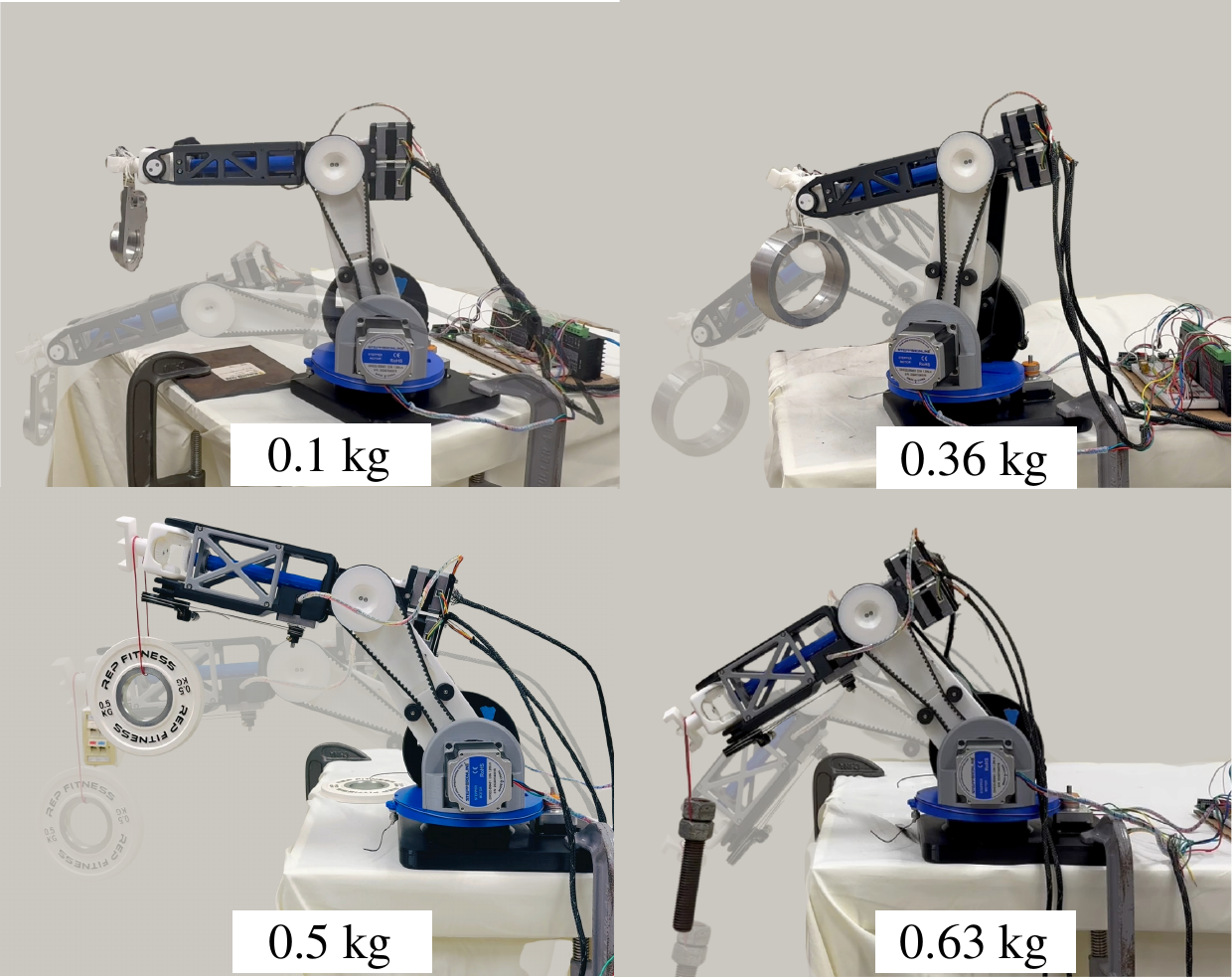}
    \caption{\textbf{Payload testing.} Forte demonstrated the capability to perform motion with payloads upto $0.63~\si{\kilogram}$.}
    \label{fig:exp2}
\end{figure}

To assess maximum payload capacity, incremental weights were added to the end-effector until performance degradation was observed. Tests were conducted with payloads of $0.1~\si{\kilogram}$, $0.36~\si{\kilogram}$, $0.5~\si{\kilogram}$, and $0.63~\si{\kilogram}$, as shown in Fig.~\ref{fig:exp2}. The system maintained stable motion and positional accuracy up to $0.63~\si{\kilogram}$. Beyond this point, the stepper motor began missing steps and could no longer generate sufficient torque to complete the motion. Future enhancements, such as improved transmission efficiency, could further increase the payload capacity closer to the theoretical value of $0.96~\si{\kilogram}$ without compromising precision.

\section{Conclusion}
In this work, we presented Forte, a fully 3D-printable, 6-DoF robotic arm that balances performance, accessibility, and cost. By leveraging a capstan-based cable drive system, timing belts, and optimized 3D-printed structures, Forte delivers sub-millimeter repeatability and supports payloads up to $0.63~\si{\kilogram}$, all at a material cost under \$215.

This level of precision was achieved through the use of high reduction ratio mechanisms, which provide both accuracy and strength, without relying on expensive position sensors or advanced control systems. Our approach prioritizes mechanical design to achieve repeatable accuracy, minimizing the need for costly feedback components. 
Experimental validation confirmed Forte’s mechanical precision, structural robustness, and suitability for educational, research, and prototyping applications.  Furthermore, Forte’s architecture holds promise for integration into humanoid platforms—particularly for upper-limb roles where low-cost, lightweight, and compact actuation is critical. Its modularity, high torque-to-weight ratio, and ability to replicate complex joint configurations make it a strong candidate for future use in affordable humanoid systems.
Forte not only extends the capabilities of low-cost robotic arms but also provides an accessible platform to foster innovation and hands-on learning in robotics—free from the constraints of expensive hardware.


\sloppy
\bibliography{citations}
\end{document}